\pdfoutput=1

\documentclass[11pt]{article}

\usepackage{authblk}

\usepackage{acl}

\usepackage{times}
\usepackage{latexsym}
\usepackage{graphicx}
\usepackage{xspace}

\usepackage[T1]{fontenc}

\usepackage[utf8]{inputenc}

\usepackage{microtype}

\newcommand{\retrieverp}{$\textbf{enc}_P$}
\newcommand{\retrievere}{$\textbf{enc}_E$}
\newcommand{\model}{\textsc{FusionED}}

\title{Entity Disambiguation via Fusion Entity Decoding}

\author{
    \textbf{Junxiong Wang}\textsuperscript{1}\thanks{\ \ Work done as an intern at Apple.}, 
    \textbf{Ali Mousavi}\textsuperscript{2}, 
    \textbf{Omar Attia}\textsuperscript{2}, 
    \textbf{Ronak Pradeep}\textsuperscript{3}\thanks{\ \ Work done as an intern at Apple.}, \\
    \vspace{2ex}
    \textbf{Saloni Potdar}\textsuperscript{2}, 
    \textbf{Alexander M. Rush}\textsuperscript{1}, 
    \textbf{Umar Farooq Minhas}\textsuperscript{2}, 
    \textbf{Yunyao Li}\textsuperscript{4} \thanks{\ \ Work done while at Apple.}\\
    \textsuperscript{1}Cornell University, 
    \textsuperscript{2}Apple Inc, 
    \textsuperscript{3}University of Waterloo,
    \textsuperscript{4}Adobe \\
    \small{junxiong@cs.cornell.edu, \{amousavi, oattia, s\_potdar, ufminhas\}@apple.com, yunyaol@adobe.com}
}

\begin{document}

\maketitle

\begin{abstract}

Entity disambiguation (ED), which links the mentions of ambiguous entities to their referent entities in a knowledge base, serves as a core component in entity linking (EL). Existing generative approaches demonstrate improved accuracy compared to classification approaches under the standardized ZELDA benchmark. Nevertheless, generative approaches suffer from the need for large-scale pre-training and inefficient generation. Most importantly, entity descriptions, which could contain crucial information to 
distinguish similar entities from each other, are often overlooked.
We propose an encoder-decoder model to disambiguate entities with more detailed entity descriptions. Given text and candidate entities, the encoder learns interactions between the text and each candidate entity, producing representations for each entity candidate. The decoder then fuses the representations of entity candidates together and selects the correct entity.
Our experiments, conducted on various entity disambiguation benchmarks, demonstrate the strong and robust performance of this model, particularly +1.5\% in the ZELDA benchmark compared with GENRE. Furthermore, we integrate this approach into the retrieval/reader framework and observe +1.5\% improvements in end-to-end entity linking in the GERBIL benchmark compared with EntQA.

\end{abstract}

\section{Introduction}

Entity linking (EL)  extracts references (a.k.a. mentions) to entities within a document and associates these mentions with their corresponding entries in a knowledge base (KB). EL is a fundamental component in automatic text comprehension, with various practical applications such as question answering, text analysis, recommender systems, semantic search, and information retrieval.

As the most critical component of EL workflows, entity disambiguation (ED) aims to select the correct entity from a set of candidate entities, given textual references. For instance, the entity mention `Bert' may stand for `the famous language model' \citep{devlin2018bert} or `the golden yellow Muppet character' depending on the given context. Therefore, models need to understand context to disambiguate entities correctly.

Owing to its practical significance in the industry and the latest developments in utilizing pre-trained language models \citep{devlin2018bert, lewis2020bart, liu2019roberta, raffel2020exploring}, various approaches for entity disambiguation have been introduced in recent years. Primarily, existing methods can be categorized into two styles: classification approaches \citep{logeswaran2019zero, yamada2022global, fevry2020empirical} or generative approaches \citep{de2020autoregressive}. Classification approaches such as \citep{yamada2022global} predict the masked entity titles while generative approaches such as \citep{de2020autoregressive} directly decode entity titles.

The recently proposed ZELDA benchmark \citep{milich2023zelda} standardizes the experimental setup (consistent training data, entity vocabulary, and candidate lists) and shows that generative approaches such as \citep{de2020autoregressive} have significantly stronger performance under this experimental setup.

However, \citet{zhang2021entqa} argues that generative approaches require large scale pre-training. In particular, \citet{de2020autoregressive} critically relies on a prefix tree (also known as a trie) derived from Wikipedia to constrain the beam search in order to produce a valid entity title in a given knowledge base (KB), which might be inefficient memory-wise. In addition, since it directly generates a valid entity without reading their descriptions, crucial information in the descriptions might be ignored. Therefore, disentangling significantly similar entities proves challenging with this method \citep{milich2023zelda}.

To better disentangle similar entities, in this paper we propose an encoder-decoder model that decodes entities by utilizing their descriptions. Our approach is mainly inspired by a recent work on question answering \citep{izacard2021leveraging}. In particular, we make the following contributions:
We summarize our contributions in the following:
\begin{itemize}
    \item We propose a new ED approach, using an encoder-decoder model. Given text and entity candidates, the encoder learns the interactions between the text and each entity candidate, generating representations for each candidate. Subsequently, the decoder fuses these candidate entity representations and generates correct entities. At inference, instead of relying on a constrained beam search, it only needs simple greedy decoding.

    \item We follow the standard evaluation practice (ensuring consistent knowledge base, training corpus and entity candidate lists) and rigorously evaluate this approach in several ED benchmarks \cite{milich2023zelda} and show its strong and robust performance.

    \item We integrate our approach into an end-to-end entity linking pipeline and show large improvements compared with the current state-of-the-art in GERBIL \cite{usbeck2015gerbil} benchmark. To the best of our knowledge, our approach is the first retrieval-augmented generation approach in EL.

    \item We propose retrieval augmented entity linking using Large Language Models (LLMs), e.g., GPT-4 and evaluate it in GERBIL \cite{usbeck2015gerbil} benchmark. Our results show that with augmented entity retrieval, GPT-4 outperforms the current SoTA on some datasets but in general, it underperforms compared to fine-tuning-based approaches.
\end{itemize}

Our approach outperforms strongest ED baselines \citep{de2020autoregressive, fevry2020empirical,yamada2022global} on ZELDA benchmark and EL baselines \citep{de2020autoregressive,zhang2021entqa,shavarani2023spel} on GEBIL benchmark \citep{usbeck2015gerbil}.

\section{Related Work}

\paragraph{Entity Disambiguation.}

Existing ED approaches typically fall into two main categories: classification approaches and generative approaches.

For classification approaches, LUKE \citep{yamada2022global} and FEVRY \citep{fevry2020empirical} are two of the most well-known approaches due to their strong performance. LUKE is based on masked entity prediction. During the pre-training, LUKE combines input text and ground-truth entities as input tokens. Then, it randomly masks entities from those ground-truth entities and predict those masked entities by leveraging both the input text and those unmasked entities. Their model is trained on a large entity-annotated corpus obtained from Wikipedia and achieves the current SoTA in several ED benchmark datasets.

For generative approaches, GENRE \citep{de2020autoregressive} uses BART weights from \citep{lewis2020bart} and is trained on a Wikipedia corpus, learning to generate entity names in an autoregressive manner, conditioned on the provided context. At inference, GENRE employs a constrained beam search strategy that forces each generated name to be in a predefined entity set.

Conventionally, ED methods are evaluated on six datasets, MSNBC, AQUAINT, ACE2004, WNED-CWEB (CWEB) and WNED-WIKI (WIKI) \citep{gabrilovich,guo2018robust}. Nevertheless, as shown in \citet{milich2023zelda}, those different ED methods use significantly different amounts of training data (ranging from 2 to 20 million annotated text) obtained with diverse sampling methodologies and enhanced weak labels \citep{orr2020bootleg,broscheit2020investigating}, and completely different knowledge bases (ranging from few thousands to over 6 million) from different sources, YAGO \citep{suchanek2007yago} or KILT \citep{petroni2021kilt} and different candidate lists \citep{hoffart2011robust,pershina2015personalized}. Thus, comparing various approaches is highly challenging. It is impossible to conclude which approach performs best \citep{milich2023zelda}.

ZELDA \citep{milich2023zelda} benchmark is proposed to unify the training data set, entity vocabulary, and candidate lists to facilitate direct comparability of ED approaches. For this reason, we compare our approach with SoTA approaches on ZELDA benchmark. Our experiment is rigorously conducted using the same training data, entity vocabulary, and candidate lists without additional information from Wikipedia or using weak labels.

\paragraph{Entity Linking.}

Different from ED, the key challenge of EL is its significantly large search space. A system can potentially generate any subset of conceivable spans in the document, each of which could correspond to an entity in a large KB, typically containing millions of entities. To manage this overwhelming scale, existing approaches break down EL into two stage tasks: mention detection (MD) and entity disambiguation (ED). These tasks are often tackled with varying degrees of independence.

In most of these approaches, the sequence of subproblems is consistent: first, the system identifies possible entity mentions, and then it links these mentions to specific entries in the given knowledge base. This MD$\rightarrow$ED classic pipeline is utilized in most methods. They either assume that mentions are provided in advance, following the example of \citet{gupta2017entity} or take a different route by employing readily available entity recognition systems to first identify mentions and then disambiguate them through the ED process, as evidenced in the works of \citet{hoffart2011robust,li2020efficient}. Furthermore, some research \citep{de2020autoregressive} trains an end-to-end autoregressive model that jointly performs MD$\rightarrow$ED by beam search.

Recently, \citet{zhang2021entqa} has shown that the classic MD $\rightarrow$ ED approach suffers from identifying mentions without prior knowledge of their corresponding entities, which is unnatural and challenging. To fix this problem, the authors flip the order of MD and ED, and propose an ED $\rightarrow$ MD pipeline. Their key observation is that finding relevant candidate entities is easy without the knowledge of their specific mentions. Their ED $\rightarrow$ MD approach achieves SoTA results on the in-domain AIDA-CoNLL dataset \citep{hoffart2011robust} and GERBIL benchmark \citep{usbeck2015gerbil}. Although their retriever (select top-$k$ candidate entities) performs remarkably well, the majority of errors are attributed to their reader (which predicts the final entities and mention spans).

A recent work \citep{shavarani2023spel} proposes a structured prediction approach and achieves 88.6\% on AIDA-CoNLL test-b by using the PPRforNED \citep{pershina2015personalized} candidate list. However, \citet{yang2018collective,milich2023zelda} question this candidate list since it is unclear how candidates were pruned. The entity candidates generated by PPRforNED \citep{pershina2015personalized} were found to be well-tailored to the AIDA-CoNLL test-b evaluation dataset, with high recall and low ambiguity. Models \citep{yamada2022global, fevry2020empirical} improve significantly when using these lists instead of the more generic lists by \citep{hoffart2011robust} and \citep{ganea2017deep}, respectively. Without the handcrafted PPRforNED \citep{pershina2015personalized} candidate list, the result of AIDA-CONLL test-b in \citep{shavarani2023spel} is the same as \citep{zhang2021entqa}, 85.8\%.

As discussed in ZELDA \citep{milich2023zelda}, using additional signals makes comparison unfair and indirect. Moreover, in real world entity linking applications, additional signals such as pruned candidate lists may not be available. Therefore, same as our comparison methodology in ED, we do not bring any additional signals and aim to conduct an end-to-end direct entity linking comparison precisely by using the same training data and same knowledge base, KILT \citep{petroni2021kilt} as EntQA \citep{zhang2021entqa} and GENRE \citep{de2020autoregressive}.

\section{Model}

\begin{figure*}
    \centering
    \includegraphics[width=\textwidth]{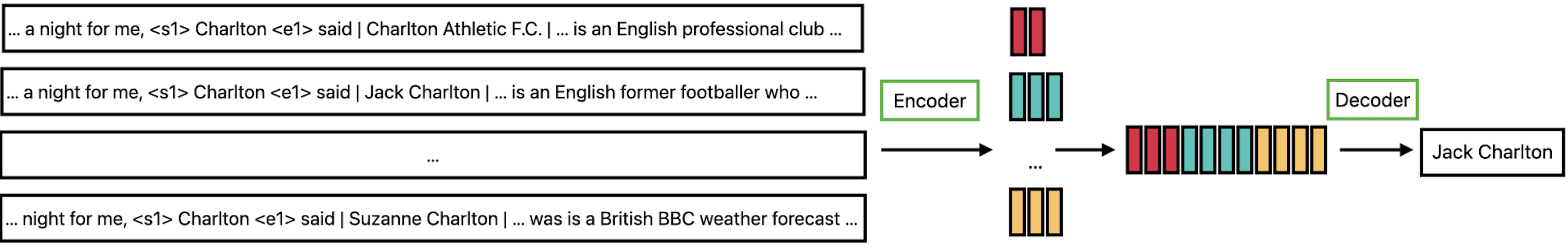}
    \caption{Pipeline of the fusion entity decoding for entity disambiguation. Given a text \textit{`DUBLIN 1996-12-07 Jack Charlton's relationship with the people of Ireland was cemented on Saturday when the Englishman was officially declared one of their own. (few sentences are abbreviated here)  That is why this is so emotional a night for me , \texttt{<s1>} Charlton \texttt{<e1>} said'}. Follow \citep{de2020autoregressive}, we add special tokens \texttt{<s1>} and \texttt{<e1>} to denote the corresponding mention to disambiguate. Given candidate entities `Charlton Athletic F.C.', `Jack Charlton', `Bobby Charlton', `Suzanne Charlton' from KB, we concatenate text with each entity candidate, including its entity title and its description. The Encoder learns interactions between the text and each entity candidate and produces suitable representations for each entity candidate; decoder concatenates those representations and selects the correct entity.
    }
    \label{fig:ed}
\end{figure*}

\subsection{Entity Disambiguation}

We formalize the ED task as follows. Given a set of candidate entities denoted as $\mathcal{E}$ in a Knowledge Base (KB), and an input text $D$ with a single mention flagged with two special start token and end token, the goal is to find the proper entity $e\in \mathcal{E}$ that corresponds to the mention in $D$.

In Figure~\ref{fig:ed}, we show an example of entity disambiguation. Given a text with annotated mention that represents what we want to disambiguate, we add special tokens \texttt{<s1>} and \texttt{<e1>} before and after the mention to denote the corresponding mention that we want to disambiguate. We concatenate input text with information from each entity candidate including entity title and entity description, and feed it into the encoder model to form an entity representation and the decoder model takes the fused entity representations from all those candidates to generate the correct entity name.

\subsection{Entity Linking} 

We formalize the EL task as follows. Given a set of entities denoted as $\mathcal{E}$ in a Knowledge Base (KB), and an input document $D$, the objective is to identify every entity $e \in \mathcal{E}$ along with a mention $m$ such that $m \in D$ and $m$ links to $e$. Typically, the length of $D$ varies from few words (e.g., short queries) to few thousands of words (e.g., news). To handle long document entity linking, previous research \citep{zhang2021entqa} typically segments each document $D$ into sentence chunks. For each sentence chunk $p$, most approaches \citep{hoffart2011robust,li2020efficient} commonly break down the task of EL for a sentence chunk $p$ into two main components: mention detection (MD) and entity disambiguation (ED), and first extract mentions from passages (MD) and then link to entities (ED).

\begin{figure}[h]
    \centering
    \includegraphics[width=0.48\textwidth]{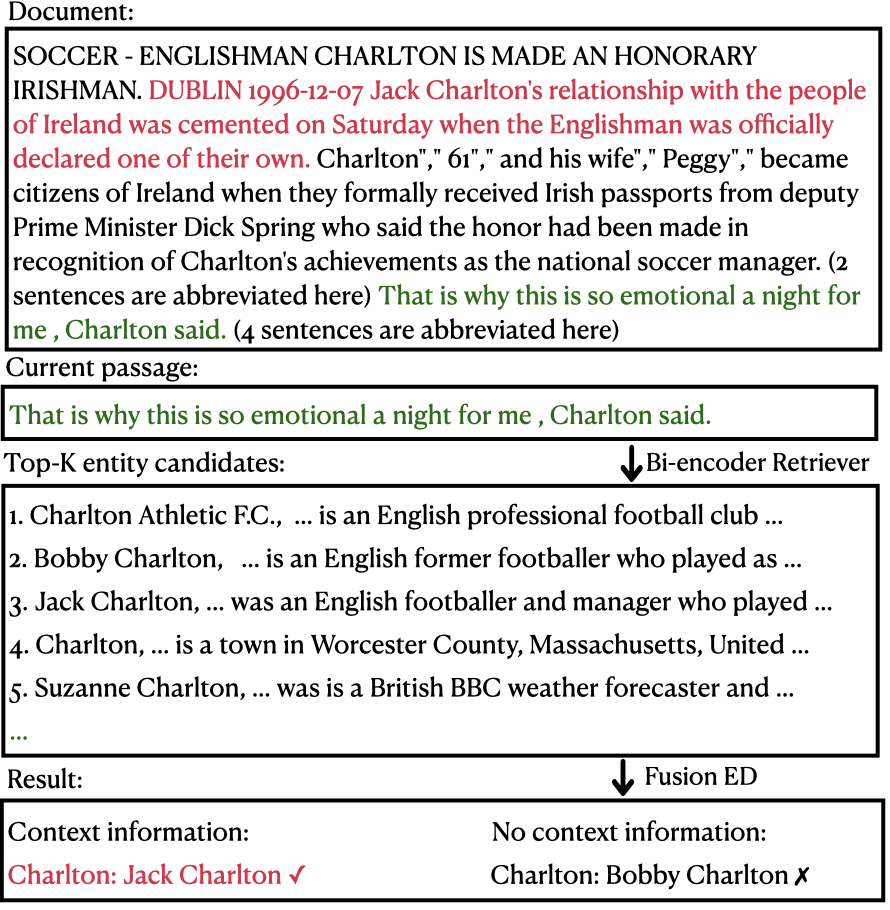}
    \caption{Example of document level entity linking from AIDA test. Given a document, \model{} splits it into smaller passage chunks. Given the current passage `That is why this is so emotional a night for me, Charlton said.', the bi-encoder entity retrieval picks up top 100 entity candidates, e.g., `Charlton Athletic F.C.', `Bobby Charlton', `Jack Charlton'. \model{} then decodes linked entities and mentions using entity candidate lists.
    }
    \label{fig:el_example}
\end{figure}

\citet{zhang2021entqa} introduce a different two-stage process, instead of first identifying mentions and then link them entities, it first retrieve top-k candidate entities, followed by the reader’s task of picking up the accurate entities along with predicting their associated mention spans. Figure~\ref{fig:el_example} illustrates an instance of end-to-end EL employing the retrieval-plus-reader approach. Our approach follow this pipeline.

\subsubsection{Bi-encoder EL Retrieval}

\paragraph{Entity Embedding.}
Following \citep{wu2019scalable}, we represent an entity $e$ as a combination of its title and description using the format: \texttt{[CLS]} \texttt{title($e$)} \texttt{[ENT]} \texttt{description($e$)} \texttt{[SEP]}. \texttt{[ENT]} is a special token to separate the entity title and description representation. For Wikipedia entities, we consider up to 128 tokens for their descriptions. We use an encoder \textbf{enc$_E$} to produce an embedding for an entity $e$.

\paragraph{Passage Embedding.}
For each passages $p$ with its document topics $t$, we also concatenate those information using the following format: \texttt{[CLS]} $p$ \texttt{[SEP]} $t$ \texttt{[SEP]}. We use another encoder \textbf{enc$_P$} to produce an embedding for a passage $p$.

\paragraph{Training.}
The score of an entity $e$ and a passage $p$ is given as
$s(e, p) = \textbf{enc}_E(e)^{\top} \textbf{enc}_P({p})$. 
Same as \citep{zhang2021entqa}, we train the retriever using a multi-label variant of noise contrastive estimation (NCE) \citep{zhang2021understanding}.

\subsubsection{Fusion EL Reader}
\label{sec:reader}

We use a similar architecture to the one we used for ED (Figure \ref{fig:ed}), while the model generates both entity names and mentions instead of only generating entity names as this was the case in ED.

Given a passage chunk $p$ along with its truncated original document $D$, the retrieval retrieves the top-$k$ candidate entities $e_1, \cdots, e_k$. Then, for each retrieved candidate entity $e_i$, we concatenate the document $D$, the current passage chunk $p$, the entity title of $e_i$, and the entity description of $e_i$. We add special tokens \texttt{<extra\_id\_0>, <extra\_id\_1>, <extra\_id\_2>, <extra\_id\_3>} before the document, the current passage chunk, the entity name, and the entity description, respectively. The input format becomes \texttt{<extra\_id\_0> D <extra\_id\_1> p <extra\_id\_2> title($e_i$) <extra\_id\_3> description($e_i$)}.

The encoder independently processes input data for each entity candidate $e_i$ and then merges the resulting representations from all the candidates. Finally, the decoder performs the attention over the merged representations of all the retrieved entities. If no candidate entities are linked, the decoder output an empty string. Otherwise, for each linked entity $e_i$, it outputs \texttt{$e_i$ 
 <extra\_id\_4> $m_{i1}, \cdots, m_{in}$} where $m_{i1}, \cdots, m_{in}$ are all mentions from $p$ which links to $e_i$. Finally, we use a special token \texttt{<extra\_id\_5>} to split the decoding output from each entity $e_i$. Therefore, the final output sting is 
\texttt{$e_1$ <extra\_id\_4> $m_{11}, \cdots m_{1n}$ <extra\_id\_5> $e_2$ <extra\_id\_4> $m_{21}, \cdots m_{2n}$ <extra\_id\_5> $\cdots$ $e_i$ <extra\_id\_4> $m_{i1}, \cdots,  m_{in}$}.

\section{Experiment}

We conduct extensive experiments to demonstrate the performance of our proposed approach (\model{}) over 20 datasets, addressing both single-entity disambiguation and end-to-end entity linking. The goal of our experiments is to facilitate a direct comparison, illustrating that under identical conditions (without incorporating extra training data or taking additional signals into account), our approach outperforms the current SoTA.

\begin{table*}[ht]
\resizebox{0.98\textwidth}{!}{
\begin{tabular}{l p{12mm} p{18mm} p{18mm} p{18mm} p{18mm} p{18mm} p{18mm} p{18mm} p{18mm} c}
\hline
Method    & AIDA-B  & TWEEKI & REDDIT-POSTS & REDDIT-COMM   & WNED-CWEB  & WNED-WIKI & SLINKS-TAIL & SLINKS-SHADOW & SLINKS-TOP & AVG \\
\hline
\textit{Baselines} & & & & & & & & & & \\
CL-RECALL & 91.1 & 94.0 & 98.4 & 98.3 & 92.4 & 98.8 & 98.8 & 56.7 & 73.1 & 89.1 \\
\hline
\textit{Classification} & & & & & & & & & & \\
FEVRY{\textsubscript{\tiny ALL}} & 79.2 & 71.8 & 88.5 & 84.1 & 68.0 & 84.3 & 63.8 & \textbf{43.4} & 53.1 & 70.7 \\
FEVRY{\textsubscript{\tiny CL}}   & 79.5 & 76.9 & 89.0 & 86.5 & 70.3 & 84.5 & 87.6 & 31.9 & 47.7 & 72.7 \\
LUKEP{\textsubscript{\tiny PRE}}  & 79.3 & 73.8 & 76.1 & 69.9 & 66.8 & 68.4 & 97.7 & 20.4 & 50.8 & 67.0 \\
LUKEP{\textsubscript{\tiny FT}} & \textbf{81.2} & 77.9 & 81.5 & 78.5 & 70.3 & 76.5 & 98.0 & 22.5 & 51.8 & 71.0 \\
\hline
\textit{Generative} & & & & & & & & & & \\
GENRE{\textsubscript{\tiny ALL}} & 72.4 & 75.9 & 88.8 & 83.9 & 66.5 & 85.2 & 95.3 & 38.7 & 43.5 & 72.2 \\
GENRE{\textsubscript{\tiny CL}}  & 78.6 & 80.1 & 92.8 & 91.5 & \textbf{73.6} & 88.4 & \textbf{99.6} & 37.3 & 52.8 & 77.2 \\
\hline
\model{}                      & \underline{80.1} & \textbf{81.4} & \textbf{93.9} & \textbf{92.3} & \textbf{73.6} & \textbf{89.0} & \underline{98.3} & \underline{41.5} & \textbf{57.9} & \textbf{78.7} \\
\hline
\end{tabular}
}
\caption{Comparison between FusionED with both classification or generative based SoTA in ZELDA Benchmark \citep{milich2023zelda}. Baselines number are taken from \citep{milich2023zelda}. We emphasize the leading model by formatting it in \textbf{bold} and the second-best model by using an \underline{underline} for each dataset. CL-RECALL represents the recall of the candidate list in ZELDA, indicating the highest possible accuracy using its candidate list.}
\label{tab:zelda}
\end{table*}

\subsection{Entity Disambiguation} 

\paragraph{Setup.}We follow the experiment setup in ZELDA benchmark \citep{milich2023zelda}, using their training data, entity vocabulary and the more generic candidate list. We initialize the weights of our model using FLAN-T5-base \citep{chung2022scaling} 220M to match the number of parameters of SoTA models (274M for LUKE \citep{yamada2022global} and FEVRY \citep{fevry2020empirical}, 178M for GENRE \citep{de2020autoregressive}). We train the model for 60k steps with a learning rate 0.0001 using Adam optimizer \citep{KingBa15}, with a batch size of 12 on 12 NVIDIA Tesla V100 32GB. 

Given a context with a mention, we consider approximately 250 tokens \footnote{ZELDA \citep{milich2023zelda} benchmark considers 500 chars to the left and 500 chars to the right of each mention. We assume that each token's length is on average and approximately equal to 4 English characters, then it results in using 250 tokens around the mention. } surrounding the annotated mention. For each entity candidate, we concatenate the entity name, a special token, and the entity description, truncating to a maximum of 140 tokens. Then, for each context, we utilize the candidate list from the benchmark \citep{milich2023zelda}. We only consider the top 200 entity candidates from this list. We evaluate checkpoints every 2000 steps for the last 8000 steps in AIDA-B, selecting the best checkpoint.


\paragraph{Datasets.} At inference, we evaluate the model using greedy decoding on 9 datasets: AIDA-B \citep{hoffart2011robust}, TWEEKI \citep{reddit2021botzer}, REDDIT-POSTS and REDDIT-COMMENTS \citep{reddit2021botzer}, WNED-WIKI and WNED-CWEB \citep{guo2018robust}, SLINKS-TOP and SLINKS-SHADOW and SLINKS-TAIL \citep{provatorova-etal-2021-robustness}. These datasets are collected from diverse sources: news (AIDA-B), annotated tweets (TWEEKI), top-scoring Reddit posts and comments (REDDIT-POSTS and REDDIT-COMMENTS), Wikipedia articles (WNED-WIKI and WNED-CWEB). In particular, \citep{provatorova-etal-2021-robustness} categorizes entities into three cases based on their appearance frequency in Wikipedia: SLINKS-TOP, where the ground truth entity is the most frequent; SLINKS-SHADOW, where a more popular entity overshadows the correct disambiguation; and SLINKS-TAIL, for rare long-tail entities.

\paragraph{Baselines.} We examine two methods presented in \citep{fevry2020empirical} using a candidate list (FEVRY{\textsubscript{\tiny CL}}) and without any restriction on the search space (FEVRY{\textsubscript{\tiny ALL}}). Additionally, for one of the ED SoTA approaches LUKE \citep{yamada2022global}, we present results of two models LUKEP{\textsubscript{\tiny PRE}} and LUKEP{\textsubscript{\tiny FT}} on ZELDA \citep{milich2023zelda} benchmark.

GENRE \citep{de2020autoregressive} employs a prefix tree derived from all entity titles in the KB to restrict the generation process. While GENRE does not utilize candidate lists during training, in inference the prefix tree can be generated using the candidate lists GENRE{\textsubscript{\tiny CL}} or without candidate lists GENRE{\textsubscript{\tiny ALL}}.

We also list CL-RECALL, which is the recall of the candidate list in ZELDA. It reflects the best possible accuracy if we always select the correct entity from the candidate list.

\paragraph{Experimental Results.}

Table~\ref{tab:zelda} reports the accuracy of \model{} compared with SoTA models. Clearly, \model{} achieves the highest performance across six datasets and secures the second position in three datasets. According to Table~\ref{tab:zelda} and as it was previously pointed out by \citep{milich2023zelda}, GENRE shows significantly better performance over classification-based baselines. However, it struggles to disambiguate entities in SLINKS-TOP and SLINKS-SHADOW. One possible interpretation is that it never uses any entity description to disambiguate entities with a similar title. Thus, it favors decoding into the most prominent case where the generated entity title will be most similar to the mention text.

It is worth mentioning that \model{} demonstrates an over +4 point accuracy improvement compared to GENRE on SLINKS-TOP and SLINKS-SHADOW datasets. These datasets involve ambiguous entities with similar titles. Incorporating information from entity descriptions is a prominent reason for \model{}'s enhanced performance.
\begin{table*}[ht]
\resizebox{\textwidth}{!}{
\begin{tabular}{lcccccccc}
\hline
Method & 1 & [1 - 0.9] & [0.9 - 0.8] & [0.8 - 0.7] & [0.7 - 0.6] & [0.6 - 0.5] & [0.5 - 0.4] & [0.4 - 0.3] \\
\hline
CL-RECALL & 99.7 & 97.2 & 99.2 & 98.3 & 98.3 & 99.1 & 98.8 & 99.6 \\
\hline
FEVRY{\textsubscript{\tiny CL}} & 94.8 & 92.2 & 88.8 & \underline{87.2} & 84.1 & 80.0 & 76.0 & 72.2 \\
LUKEP{\textsubscript{\tiny FT}} & 91.5 & 90.4 & 86.3 & 80.3 & 77.8 & 73.8 & 62.2 & 56.2 \\
GENRE{\textsubscript{\tiny CL}} & \textbf{97.1} & \textbf{94.2} & \textbf{91.2} & 85.6 & 87.8 & 86.9 & \underline{86.9} & \underline{79.7} \\
\hline
\model{} & \underline{96.4} & \underline{92.4} & \underline{90.8} & \textbf{87.5} & \underline{86.1} & \textbf{88.1} & \textbf{87.1} & \textbf{85.0} \\
\hline
\end{tabular}
}
\caption{Accuracy across various difficulty brackets was assessed for different approaches in the WNED-WIKI dataset. [0.4 - 0.3] is the most difficult bracket while 1 is the easiest. We emphasize the leading model by highlighting it in bold and denote the runner-up with an underline for each bracket. Our model shows the best performance across most different brackets, suggesting that using entity descriptions can help disambiguate closed entities in most challenging tests.}
\label{tab:zelda_ablation}
\end{table*}

Table~\ref{tab:zelda_ablation} shows the accuracy of different approaches across various difficulty brackets in the WNED-WIKI dataset, introduced in \citep{guo2018robust}. They propose a baseline method PRIOR by selecting the entity with the highest prior probability, denoted as $prior(m, e)$, for a given mention $m$. This prior probability is precomputed using all annotated mention-entity pairs from Web-scale and Wikipedia corpora. PRIOR serves as a proxy to assess the difficulty of a mention. They further normalize the probability of the ground truth entity given mention. Based on this normalized value, they categorize difficulty into eight brackets. Specifically, if the probability for the corresponding ground truth entity of a mention is low, indicating increased ambiguity across the entire Web and Wikipedia corpora, the mention is considered more difficult. [0.4 - 0.3] represents the most difficult test cases while 1 represents the easiest ones. Our model has the highest accuracy across most different brackets (+5\% in [0.4 - 0.3]), suggesting that using entity descriptions can help disambiguate closed entities in most challenging test cases. 

\subsection{Entity Linking}

\begin{table*}[ht]
\resizebox{\textwidth}{!}{%
\begin{tabular}{lcccccccc|c}
\hline
          & In-domain  &  \multicolumn{7}{c}{Out-of-domain} &  \\
Method     & AIDA-B  & MSNBC & Der   & K50   & R128  & R500   & OKE15 & OKE16  & AVG     \\
\hline
\citet{hoffart2011robust}      & 72.8 & 65.1 & 32.6 & 55.4 & 46.4 & \textbf{42.4}  & \textbf{63.1}  & 0  & 47.2   \\
\citet{steinmetz2013semantic}  & 42.3 & 30.9 & 26.5 & 46.8 & 18.1 & 20.5  & 46.2  & 46.4  & 34.7  \\
\citet{moro2014entity}         & 48.5 & 39.7 & 29.8 & 55.9 & 23.0 & 29.1  & 41.9  & 37.7  & 38.2  \\
\citet{kolitsas2018end}        & 82.4 & 72.4 & 34.1 & 35.2 & 50.3 & 38.2  & \underline{61.9}  & 52.7  & 53.4   \\
\citet{broscheit2020invest}    & 79.3 & -    & -    & -    & -    & -     & -     & -     &    \\
\citet{martins2019joint}       & 81.9 & -    & -    & -    & -    & -     & -     & -     &    \\
\citet{van2020rel}             & 80.5 & 72.4 & 41.1   & 50.7 & 49.9   & 35.0 & \textbf{63.1} & \textbf{58.3}  & 56.4   \\
\citet{de2020autoregressive}   & 83.7 & \textbf{73.7} & 54.1 & 60.7  & 46.7 & 40.3 & 56.1  & 50.0   & 58.2   \\
\citet{de2021highly}           & 85.5    & -    & -    & -    & -    & -     & -     & -     &    \\
\citet{zhang2021entqa}         & \underline{85.8} & 72.1 & 52.9   & 64.5 & \textbf{54.1} & \underline{41.9}  & 61.1  & 51.3   & \underline{60.5}   \\
\citet{shavarani2023spel}      & \underline{85.8} & 63.1 & \textbf{59.1}   & 53.7 & 47.1 & 44.4 & 59.5 & 56.6 & 58.7 \\
GPT-4 (zero-shot) \citet{shavarani2023spel} & 54.1 & -    & -    & -    & -    & -     & -     & -     &    \\
GPT-4 + retrieval (zero-shot) & 58.4 & 42.4 & 40.1 & \textbf{69.0} & 35.1 & 29.4 & 58.3 & 53.1 & 48.3 \\
GPT-4 + retrieval (zero-shot)* & 59.1 & 42.5 & 41.0 & 67.6 & 36.4 & 30.1 & 58.4 & 53.0 & 48.5 \\
\hline
\model{}                       & \textbf{86.5} & \underline{73.6} & \underline{56.8} & \underline{65.1} & \underline{53.1} & 41.6  & 62.3 & \underline{56.6}  & \textbf{62.0}   \\
\hline
\end{tabular}%
}
\caption{InKB Micro F1 on the GERBIL benchmark with respect to in-domain and out-of-domain test sets. We highlight the top-performing model in \textbf{bold} and the runner-up in \underline{underline} for each dataset. For \citep{shavarani2023spel}, to make a fair comparison, we use their AIDA-testb result without external additional candidate set \citep{pershina2015personalized}. For GPT-4 + retrieval (zero-shot)*, we additionally filter entities generated by the model using candidate entities obtained from entity retrieval and this slightly improve its overall performance.}
\label{tab:gerbil}
\end{table*}

\paragraph{Setup.} 
For EL, we adhere to the established convention \citep{de2020autoregressive,zhang2021entqa} by presenting the InKB Micro F1 score for both the in-domain and out-of-domain datasets. Specifically, for the in-domain scenario, we train \model{} using the AIDA-CoNLL dataset \citep{hoffart2011robust}. For the out-of-domain tests, following the same practice, we evaluate it on seven test sets: MSNBC \cite{cucerzan2007large}, Derczynski (Der) \citep{derczynski2015analysis}, KORE 50 (K50) \cite{hoffart2012kore}, N3-Reuters-128 (R128), N3-RSS-500
(R500) \citep{roder2014n3}, and OKE challenge 2015 and 2016 (OKE15 and OKE16) \citep{nuzzolese2015open}. For KB, we utilize the 2019 Wikipedia dump, as supplied within the KILT benchmark \citep{petroni2021kilt}, encompassing a total of 5.9 million entities for our knowledge base (KB).

\paragraph{Retriever Training.}

Following \citep{zhang2021entqa}, we initialize weights of both the passage encoder (\retrieverp{}) and the entity encoder (\retrievere{}) using BLINK \citep{wu2019scalable} retrievers that have been pretrained on Wikipedia hyperlinks. We also finetune retrievers using NCE objective with hard negative mining and follow the same sampling strategy as \citep{zhang2021entqa} (90\% from random sample and 10\% from hard negatives) . We reproduce their retriever by matching their top 100 recall numbers reported in their paper. We use FAISS \citep{johnson2019billion} to speed up vector similarity search.

\paragraph{Reader Training.}

We create the reader dataset by selecting the top 100 candidates from the retrieval process. For each ground truth entity, we create an entity title and mention pair. And we concatenate truncated document and those entity pairs together as discussed in section \ref{sec:reader} \footnote{EntQA \citep{zhang2021entqa} shows injecting document level information can improve model performance largely (+1.4 F1). So we use a truncated document of up to 20 tokens, which roughly corresponds to the first sentence approach in EntQA \citep{zhang2021entqa}}. 

The model is initialized with the FLAN-T5-large model \citep{chung2022scaling}. We finetune the model for 20k steps with a learning rate of 0.0001 using the Adam optimizer \citep{KingBa15}, with a batch size of 8, employing 8 NVIDIA Tesla A100 40GB GPUs. Following the approach in \citep{zhang2021entqa}, we evaluate the models every 1000 steps in AIDA and select the best checkpoint. We use a linear decay learning rate scheduler that starts at 0, warms up to the peak learning rate, and then decays back to 0. The warm-up rate is set to 1\%. 

\paragraph{Inference.}
During inference, we employ a sliding window approach to split the document into passages with a window size of 20 tokens and a stride of 10 tokens to avoid cutting off any mentions. For each split passage, we first retrieve the top 100 entity candidates using the bi-encoder, followed by a \model{} reader to decode correct entities along with their mentions. Using a sliding window approach might cause the reader to identify overlapping mentions or to disambiguate a single mention into two different entities. For overlapping mentions, we retain the longest one. And if the same mention is disambiguated into two different entities, we retain both entities.

\paragraph{Experimental Results.}

Table~\ref{tab:gerbil} shows InKB Micro F1 of \model{} compared with different entity linking systems. Clearly, \model{} achieves the best in-domain test (+0.7\% F1 for AIDA-B \citep{hoffart2011robust}) without using any handcrafted candidate list \citep{pershina2015personalized}

Overall, \model{} achieves the best averaged F1 score across the all evaluation datasets; +1.5\% over EntQA \citep{zhang2021entqa} and +2.8\% over the latest work \citep{shavarani2023spel} in EL. The reason for the lower performance on OKE15 and OKE16 \citep{nuzzolese2015open} is consistent with the observation made by \citep{de2020autoregressive}: these datasets include coreference annotations (such as pronouns and common nouns linked to entities), for which our model lacks training. In contrast, many other systems incorporate a component in their pipelines specifically designed to use these annotations.

Compared to the previous retrieval-plus-reader approach, EntQA \citep{zhang2021entqa}, \model{} improves by +1.5\% on MSNBC, +3.9\% on Der, +0.6\% on K50, and +4.7\% on OKE16. 

\subsection{Case Study: Retrieval-augmented LLMs for Entity Linking}

\begin{table}[th]
\resizebox{0.48\textwidth}{!}{%
\begin{tabular}{lcccccc}
\hline
Datasets& \multicolumn{3}{c}{GPT-4 + retrieval}  &  \multicolumn{3}{c}{\model{}}  \\
& P & R & F1 & P & R & F1  \\
\hline
AIDA-B  & 52.0 & 66.6 & 58.4 & \textbf{84.4} & \textbf{88.7} & \textbf{86.5} \\
MSNBC   & 32.6 & 60.7 & 42.4 & \textbf{75.6} & \textbf{71.7} & \textbf{73.6} \\
Der     & 29.2 & \textbf{63.9} & 40.1 & \textbf{55.2} & 58.5 & \textbf{56.8} \\
K50     & 70.3 & \textbf{67.8} & \textbf{69.0} & \textbf{72.0} & 59.4 & 65.1\\
R128    & 25.6 & \textbf{55.6} & 35.1 & \textbf{56.3} & 50.2 & \textbf{53.1} \\
R500    & 19.2 & \textbf{62.8} & 29.4 & \textbf{31.6} & 60.7 & \textbf{41.6} \\
OKE15   & 64.1 & \textbf{53.5} & 58.3 & \textbf{80.1} & 51.0 & \textbf{62.3} \\
OKE16   & 60.7 & \textbf{47.2} & 53.1 & \textbf{76.8} & 44.8 & \textbf{56.6}\\
\hline
\end{tabular}%
}
\caption{In contrast to \model{}, GPT-4 + retrieval demonstrates improved recall (R) across all datasets except AIDA-B and MSNBC, while exhibiting inferior precision (P) across all datasets.}
\label{tab:gpt4_ablation}
\end{table}

\citet{shavarani2023spel} has benchmarked LLMs for EL using the approach introduced in \citep{de2020autoregressive} where it produces a markup around the mentions followed by the linked entity name. However, the results are much worse than our approach, 54.1 vs 86.5. Although LLMs possess comprehensive knowledge about entities, they face a limitation in directly reasoning about specific Wikipedia URLs and Wikipedia names.

We conduct a preliminary study to assess the performance of retrieval-augmented prompting for linking entities using LLMs. This approach involves utilizing the same retrieval models that we described before, which are initialized using BLINK \citep{wu2019scalable} weights and fine-tuned based on AIDA \citep{hoffart2011robust}. For the reader, we replace the \model{} with GPT-4. More precisely, we provide GPT-4 with truncated documents (up to 50 tokens), input passages, and entity candidates, including entity title and entity description (up to 50 tokens). We prompt it to link entities from the candidate entity sets and identify their corresponding mentions. To the best of our knowledge, we are the first to propose retrieval-augmented LLMs for EL.

Table~\ref{tab:gpt4_ablation} presents a detailed comparison between \model{} and GPT-4 + retrieval. GPT-4 + retrieval shows better recall (R) in all datasets except AIDA-B, MSNBC, but it has lower precision (P) in all datasets. The inferior precision of GPT-4 might stem from 1) ambiguity in defining entities, where it considers instances like `Spoon',  `Pasta', `Scientist' as entities diverge from actual ground truth labels in MSNBC \citep{cucerzan2007large}; 2) linking ambiguous partial names to famous entities (e.g., in a dataset based on tweets \citep{derczynski2015analysis}, a given query is `I'm going home to Wisconsin', it links the ambiguous entity `Wisconsin' to the Wisconsin state, but it may refer to `University of Wisconsin–Madison'). Our preliminary results suggest that future research should focus on enhancing the precision of LLMs by using varied prompts to match SoTA fine-tuned models.

\section{Conclusion}

We propose an encoder-decoder model architecture to enhance the disambiguation of entities by providing more detailed descriptions. The encoder, when given text and candidate entities learns the interactions between the text and each entity candidate, generating representations for each candidate. The decoder then combines these representations to produce the correct entity. Our experiments, conducted on various entity disambiguation benchmarks, demonstrate the model's strong and robust performance. Furthermore, we integrate this approach into the retrieval/reader EL framework and observe improvements on the GERBIL benchmark compared with previous SoTA. We also propose entity retrieval-augmented large language models (LLMs) for EL. Results show that compared to \model{}, LLMs generally underperform while they demonstrate strong improvements compared to SoTA over some datasets.

\section{Limitations and Ethical Considerations}
The scope of our ED and EL models are limited to traditional Wikipedia and News datasets. We have not investigated its effectiveness in diverse domains such as biomedical research, e-commerce, and product catalogs. Furthermore, this paper focuses exclusively on the English corpus, and exploring the potential of our model in a multilingual setting would be an interesting expansion for future research. This includes investigating the advantages of projecting entity linking concepts from one language to another and employing multilingual representation learning to enhance our base model. While our retrieval-augmented LLMs exhibit notable performance improvements for certain datasets in EL, they underperform compared to the other approaches. Investigating how to enhance the performance of LLMs using different prompts further is an interesting direction for exploration.

Our models are trained using datasets comprised of existing textual collections sourced from Wikipedia and News. Recent studies have brought attention to potential societal biases ingrained in established corpora. We acknowledge the potential risk that our EL models may inherit such biases.

\section*{Acknowledgements}
We thank Simone Conia, Farima Fatahi Bayat, Yifu Qiu, Revanth Gangi Reddy, and Zoey Li for their valuable suggestions and feedback. We also appreciate the extensive discussions we had with Edouard Grave as he introduced his excellent work on fusion-in-decoder.

\bibliography{anthology,custom}
\bibliographystyle{acl_natbib}

\begin{table*}[!th]
\resizebox{\textwidth}{!}{%
\begin{tabular}{lcccccc|c}
\hline
          & In-domain  &  \multicolumn{5}{c}{Out-of-domain} &  \\
Method    & AIDA-B  & MSNBC & AQUAINT   & ACE2004   & CWEB  & WIKI & AVG \\
\hline
\citet{de2020autoregressive}   & 88.6 & 88.1 & 77.1 & 82.3  & 71.9 & 71.7 & 80.0 \\
\model{}                       & 91.7 & 92.4	& 82.0	& 87.1 & 75.8 &	78.6 & 84.6 \\
\hline
\end{tabular}%
}
\caption{InKB Micro F1 comparison of GENER and \model{} when only training in AIDA dataset and evaluate the performance on both in-domain and out-of-domain. The goal of this experiments is to provide a direct comparison.}
\label{tab:ed_ned}
\end{table*}

\newpage 

\appendix

\section{Additional Experiments on Named Entity Disambiguation Benchmark}


We also run a small ablation experiment on traditional named entity disambiguation datasets using FLAN-T5-large as base model to compare the corresponding large model. Unlike a standard benchmark, models which test on those datasets typically trained using different corpus and linked to different KB which maybe subset of YAGO \citep{suchanek2007yago} and KILT \citep{petroni2021kilt}. Reproducing those results might be a challenge due to the incomplete release of their entity vocabulary \footnote{\url{https://github.com/facebookresearch/GENRE/issues/26}} \footnote{\url{https://github.com/facebookresearch/GENRE/issues/72}}. And comparison is indirect since training datasets are different and may overlap with some test datasets used in out-of-domain evaluation \footnote{\url{https://github.com/facebookresearch/GENRE/issues/13}}.

We avoid training our model on Wikipedia datasets to prevent test data leakage. Instead, we conduct ablation experiments, training on AIDA and evaluating it in both in-domain AIDA-B and out-domain datasets such as MSNBC, AQUAINT, ACE2004, WNED-CWEB (CWEB), and WNED-WIKI (WIKI) \citep{gabrilovich, guo2018robust} to provide a direct comparison.

At the inference, we rely on the same candidate lists provided in \citep{de2020autoregressive} \footnote{\url{https://github.com/facebookresearch/GENRE/tree/main/examples_genre}}. Instead of decoding entity names, we decode the corresponding entity number in the given ordered candidate list.


Table~\ref{tab:ed_ned} presents a comparison of InKB Micro F1 results between GENER and \model{} when only trained on the AIDA dataset and evaluated in both in-domain and out-of-domain scenarios. \model{} shows much better performance compared to GENER, supporting our claim that our model does not require significant pre-training. It is worth noting that our numbers are not directly comparable with SoTA models, as those models are trained on different corpus.


\section{Entity Linking Experiments in GPT-4}

Our prompt template is as follows:

\texttt{Given a input passage and a candidate entity list (each element in this list is a pair with entity title and entity description), your task is to select entities from this list and link them to mentions which appear in given passage. For each linkage, please output the entity title and mention, separated by @\#@ on each line. You can use the truncated document as context information. passage: ... , entities: ... , document: ...}

For each passage, we first retrieve the top-100 entity candidates, then feed this passage, entity candidates, and the corresponding truncated document into this template to produce a prompt. Subsequently, we call the GPT-4-16k API to get results. Then we parse results and evaluate those in GERBIL benchmark.

\begin{table}[th]
\resizebox{0.48\textwidth}{!}{%
\begin{tabular}{lcccccc}
\hline
& \multicolumn{3}{c}{GPT-4 + retrieval}  &  \multicolumn{3}{c}{GPT-4 + retrieval*}  \\
Dataset & P & R & F1 & P & R & F1  \\
\hline
AIDA-B  & 52.0 & \textbf{66.6} & 58.4 & \textbf{53.2} & 66.5 & \textbf{59.1} \\
MSNBC   & 32.6 & \textbf{60.7} & 42.4 & \textbf{32.8} & 60.5 & \textbf{42.5} \\
Der     & 29.2 & \textbf{63.9} & 40.1 & \textbf{30.2} & \textbf{63.9} & \textbf{41.0} \\
K50     & 70.3 & \textbf{67.8} & \textbf{69.0} & \textbf{72.0} & 59.4 & 65.1\\
R128    & 25.6 & \textbf{55.6} & 35.1 & \textbf{27.2} & 55.2 & \textbf{36.4} \\
R500    & 19.2 & \textbf{62.8} & 29.4 & \textbf{20.1} & 61.0 & \textbf{30.1} \\
OKE15   & 64.1 & \textbf{53.5} & 58.3 & \textbf{64.6} & 53.3 & \textbf{58.4} \\
OKE16   & 60.7 & \textbf{47.2} & \textbf{53.1} & \textbf{61.5} & 46.5 & 53.0\\
\hline
\end{tabular}%
}
\caption{Breakdown of the score, Precision (P), Recall (R) and F1 for the GPT-4 + retrieval method.}
\label{tab:gpt4_ablation2}
\vspace{-1em}
\end{table}

Table~\ref{tab:gpt4_ablation2} presents the results of GPT-4 in the GERBIL benchmark \citep{usbeck2015gerbil}. For GPT-4 + retrieval (zero-shot)*, we additionally filter entities generated by the model using candidate entities obtained from entity retrieval and this improves its precision and slightly improve its performance over all datasets except K50 and OKE16.

\end{document}